\documentclass[10pt,twocolumn,letterpaper]{article}

\usepackage{3dv}
\usepackage{times}
\usepackage{epsfig}
\usepackage{subfig}
\usepackage{graphicx}
\usepackage{amsmath}
\usepackage{amssymb}
\usepackage{float}
\usepackage{caption}
\usepackage{units}
\usepackage{tabularx}
\usepackage{booktabs}
\usepackage{multirow}
\usepackage{verbatim}
\usepackage{enumitem}

\usepackage[nolist,nohyperlinks]{acronym}
\usepackage{microtype}

\usepackage[pagebackref=true,breaklinks=true,letterpaper=true,colorlinks,bookmarks=false]{hyperref}

\threedvfinalcopy 

\def\threedvPaperID{225} 

\ifthreedvfinal\fi

\newcommand\blfootnote[1]{%
  \begingroup
  \renewcommand\thefootnote{}\footnote{#1}%
  \addtocounter{footnote}{-1}%
  \endgroup
}

\begin{document}

\title{LCD -- Line Clustering and Description for Place Recognition}

\author{Felix Taubner$^1$, Florian Tschopp$^1$, Tonci Novkovic$^{1,2}$, Roland Siegwart$^1$, and Fadri Furrer$^{1,3}$\\
$^1$ETH Zurich, Switzerland, $^2$Gideon Brothers, $^3$incon.ai\\
{\tt\small \{ftaubner,ftschopp,rsiegwart\}@ethz.ch, tonci.novkovic@gideonbros.ai, fadri@incon.ai}

}

\maketitle

\begin{abstract}
   Current research on visual place recognition mostly focuses on aggregating local visual features of an image into a single vector representation. Therefore, high-level information such as the geometric arrangement of the features is typically lost. 
   In this paper, we introduce a novel learning-based approach to place recognition, using RGB-D cameras and line clusters as visual and geometric features.
   We state the place recognition problem as a problem of recognizing clusters of lines instead of individual patches, thus maintaining structural information. 
   In our work, line clusters are defined as lines that make up individual objects, hence our place recognition approach can be understood as object recognition.
   3D line segments are detected in RGB-D images using state-of-the-art techniques. We present a neural network architecture based on the attention mechanism for frame-wise line clustering.
   A similar neural network is used for the description of these clusters with a compact embedding of 128 floating point numbers, trained with triplet loss on training data obtained from the InteriorNet dataset.
   We show experiments on a large number of indoor scenes and compare our method with the bag-of-words image-retrieval approach using SIFT and SuperPoint features and the global descriptor NetVLAD.
   Trained only on synthetic data, our approach generalizes well to real-world data captured with Kinect sensors, while also providing information about the geometric arrangement of instances. 
\end{abstract}
\blfootnote{Code is available at \url{https://github.com/ethz-asl/lcd}}
\blfootnote{This works was partially supported by Siemens Mobility GmbH, Germany. We would like to thank Francesco Milano, Dominik Walder and Chengkun Li for their initial work.}
\blfootnote{\textcopyright 2020 IEEE. Personal use of this material is permitted. Permission from IEEE must be obtained for all other uses, in any current or future media, including reprinting/republishing this material for advertising or promotional purposes, creating new collective works, for resale or redistribution to servers or lists, or reuse of any copyrighted component of this work in other works
}
\section{Introduction}
Visual place recognition is a very active area of research in computer vision and robotics \cite{PlaceRecognitionSummary}. In order to be able to navigate efficiently in its environment, a robot needs to know where it is and recognize scenes where it has been before. In GPS-denied areas, visual place recognition can be used as an alternative.
Various current successful place recognition systems  \cite{ImproveObjectRetrieval,FABMAP,netvlad,Lost} cast place recognition as an image retrieval problem. Many of them use the \ac{bow} approach \cite{BagOfWords,ObjectRetrieval}, which aggregate visual features such as SIFT~\cite{sift} or SURF~\cite{Surf} to describe images with a unique representation vector. Places are recognized through similarity of visual appearance. 
\begin{figure}[bt!]
    \centering
        \includegraphics[width=0.32\columnwidth]{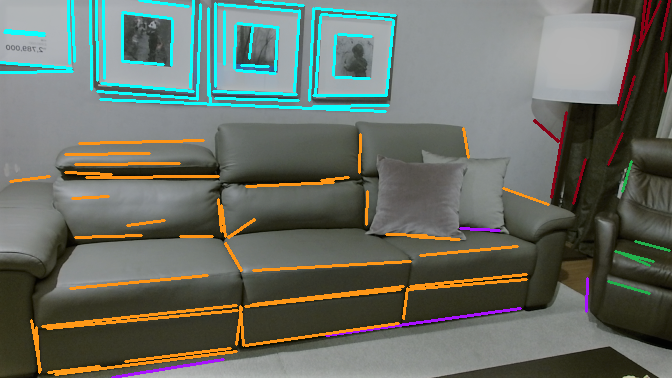}
        \includegraphics[width=0.32\columnwidth]{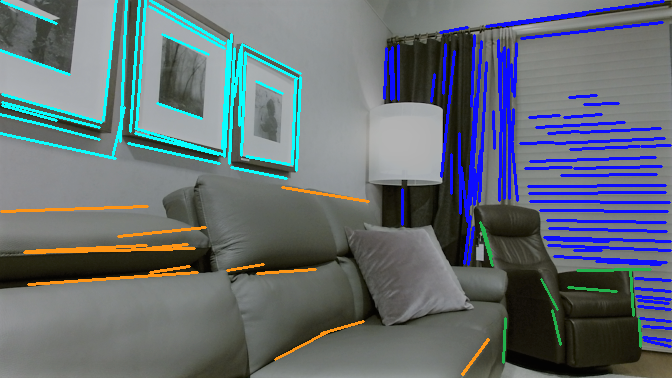}
        \includegraphics[width=0.32\columnwidth]{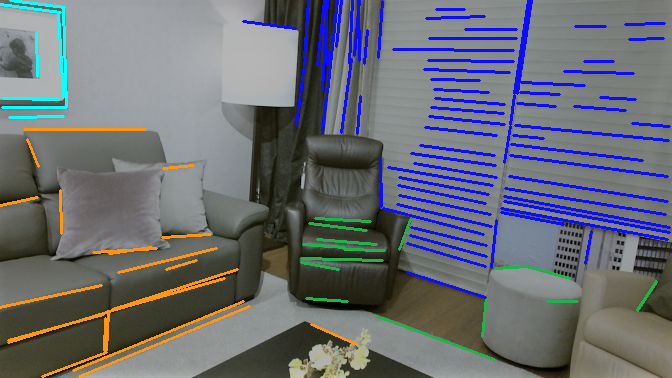}
    \caption{We present a novel place recognition method that uses clusters of lines as visual and geometric features. Lines are detected in RGB-D images and clustered using a neural network. Another neural network describes the line clusters with a compact 128-dimensional embedding. These descriptors can be matched to find the corresponding scene. The colors of the lines in the images represent different clusters, and line clusters of the same color represent correct matches under different viewpoints. }
    \label{fig:teaser}
    \vspace{-0.6cm}
\end{figure}

Recent advancements in \ac{dl} and the appearance of large-scale labelled datasets have sparked numerous novel learning-based approaches for image retrieval and place recognition. Most of these use \ac{cnn} features to describe images \cite{LocalDeepFeatures,ConvNetLandmarks}.
While these approaches show impressive results, they typically work only with similar viewpoint directions and struggle with illumination changes.

In this work, we want to incorporate also structural cues for the place recognition task. Specifically, as clear structures are often found in human made environments, we anticipate that 3D line segments can improve place recognition. We propose to use line clusters as visual features, for the following reasons:
First, lines\footnote {For brevity, the terms line segment and line are used interchangeably in this paper.} can be clustered to form structural instances which we assume to be less variant to illumination or viewpoint changes.
Second, there are plenty of lines in human made environments. Finally, lines contain more structural information than keypoints.

Typically, visually detectable lines are object edges or texture lines. These lines can be seen from most viewing angles, while not being affected by lighting changes. In~\cite{RobustPointAndLines1} and \cite{RobustPointAndLines2} the authors demonstrate the advantage of using lines as features under varying lighting conditions. In contrast, we employ a virtual camera image for description to improve viewpoint invariance.

Furthermore, lines describe object structures very well, because object contours are often made up of straight lines in human-made environments (Figure~\ref{fig:teaser}). 
We utilize the structural property of lines by describing entire clusters of line segments.
What makes up a cluster is determined by the way we train the networks. 
A human would intuitively recognize scenes by splitting it into recognizable objects, such as a certain model of a chair, window, table, etc. 
Hence, we specify line clusters as the lines that belong to certain semantic instances. We define places as a conglomeration of instances that make up a scene. Thus, the task of place recognition is cast as the problem of recognizing specific instances. Instance-level visual features provide a high-level description of the place while maintaining invariance to the displacement of objects.

Our work can be divided into four main contributions that function largely independently of each other: 
\begin{enumerate}[topsep=5pt,itemsep=-1ex,partopsep=1ex,parsep=1ex]
    \item A pipeline for robust detection and reprojection of 3D line segments in an RGB-D image
    \item A novel method to obtain viewpoint invariant visual descriptions of lines via virtual camera images
    \item An attention-mechanism based neural network for frame-wise clustering of lines at instance level
    \item A neural network for the description of line clusters with compact 128-dimensional embeddings.
\end{enumerate}

\section{Related work}
%
Visual descriptors can be categorized as global descriptors, which describe the whole image, and local descriptors, describing local patches in the image~\cite{PlaceRecognitionSummary}. 
Among the most popular hand-crafted local descriptors are SIFT, SURF and BRISK~\cite{ma2020image}. \ac{dl}-based descriptors such as \ac{sp}~\cite{SuperPoint} have been introduced recently.
Such local descriptors are typically combined using \ac{bow}~\cite{BagOfWords} to create a global image descriptor.
%
%
However, information on the geometric arrangement of features is lost.
Alternatively, global descriptors such as Gist~\cite{Gist} and VLAD~\cite{arandjelovic2013vlad} compress the information from the whole image into a distinctive descriptor.
More recent advances in image retrieval \cite{netvlad,weyand2016planet} use neural networks for descriptor extraction. 
While global descriptors are more invariant to appearance changes caused by season and lighting, they are more susceptible to variations in viewpoint~\cite{PlaceRecognitionSummary}.
With the availability of RGB-D data, point features in 3D have become more popular~\cite{3DSift, ma2020image}.
They capture more geometric data, but are susceptible to noise and incomplete 3D data~\cite{spezialetti20203d} which can be alleviated to some extent with learning approaches~\cite{Gojcic_2019_CVPR, 3Dmatch}.
%

In contrast to point features, lines can better handle illumination and viewpoint changes, are more robust in low textured environments with occlusions, convey structural information and provide complementary information~\cite{OutdoorStraightLines,verhagen2014scale,7298936}.
Typical ways to detect lines are shown in~\cite{LSD, edlines}.
LDB~\cite{zhang2013efficient} and MSLD~\cite{MSLD_line} use line surroundings for description, and are extended for scale invariance~\cite{verhagen2014scale} and further improved using neural networks~\cite{DLD_line}. 
Geometric properties of clustered nearby lines are used for description in~\cite{wang2009wide}.
%

Since global descriptors can only provide an approximate pose corresponding to the closest previously observed image, a localization approach that combines local and global visual features was proposed~\cite{sarlin2019coarse}.
A city scale localization scheme is presented in~\cite{svarm2016city}.
Semantic information can further improve localization~\cite{toft2018semantic}.
Using existing region proposal networks to detect landmarks of interest and a \ac{cnn} based visual descriptor to identify and recognize these landmarks is presented in~\cite{ConvNetLandmarks}.
In addition, RGB-D images and semantic segmentation can be used to describe the structure of object arrangements~\cite{PhysicalWords}.
3D features~\cite{ObjectDatabase} and later a combination of 2D and 3D features~\cite{furrer2020phd_thesis} enable the matching and combination of objects.
The \ac{bow} approach is used in a wide variety of applications, \eg in~\cite{IncrementalSLAM,SceneSequences,ContentBasedRetrieval} and extended to include 3D information ~\cite{StereoSequences,FABMAP2_0,FABMAP3D,SiftSurfSeasons}. %
\cite{6631111} use MSLD and a vocabulary tree to do \ac{slam}.
\cite{7298936} use line segments to localize in a map by matching to a virtual view.
\cite{Tang2019place} propose place recognition using line pairs and \ac{bow}, showing better performance than ORB.
The advantage of using line segments as visual features has been demonstrated for \ac{slam}~\cite{3DLineBasedSLAM,RobustPointAndLines1} and place recognition~\cite{OutdoorStraightLines,LineClusteringLane}.

Our work uses a hybrid between the local and global descriptor approach by detecting lines at instance-level and describing clusters of lines for place recognition.   
We propose a novel line detection method in 3D, based on LSD, that uses depth information to obtain the 3D lines and a viewpoint-invariant image of the line environments.
We use instance-level line clusters as features and a learned visual-geometric descriptor to recognize places.
No previous work demonstrated an approach that uses instance-level descriptors with structural information while taking advantage of the robustness of lines as visual features.

\section{Method}

\begin{figure*}
    \begin{center}
        \includegraphics[width=.85\linewidth]{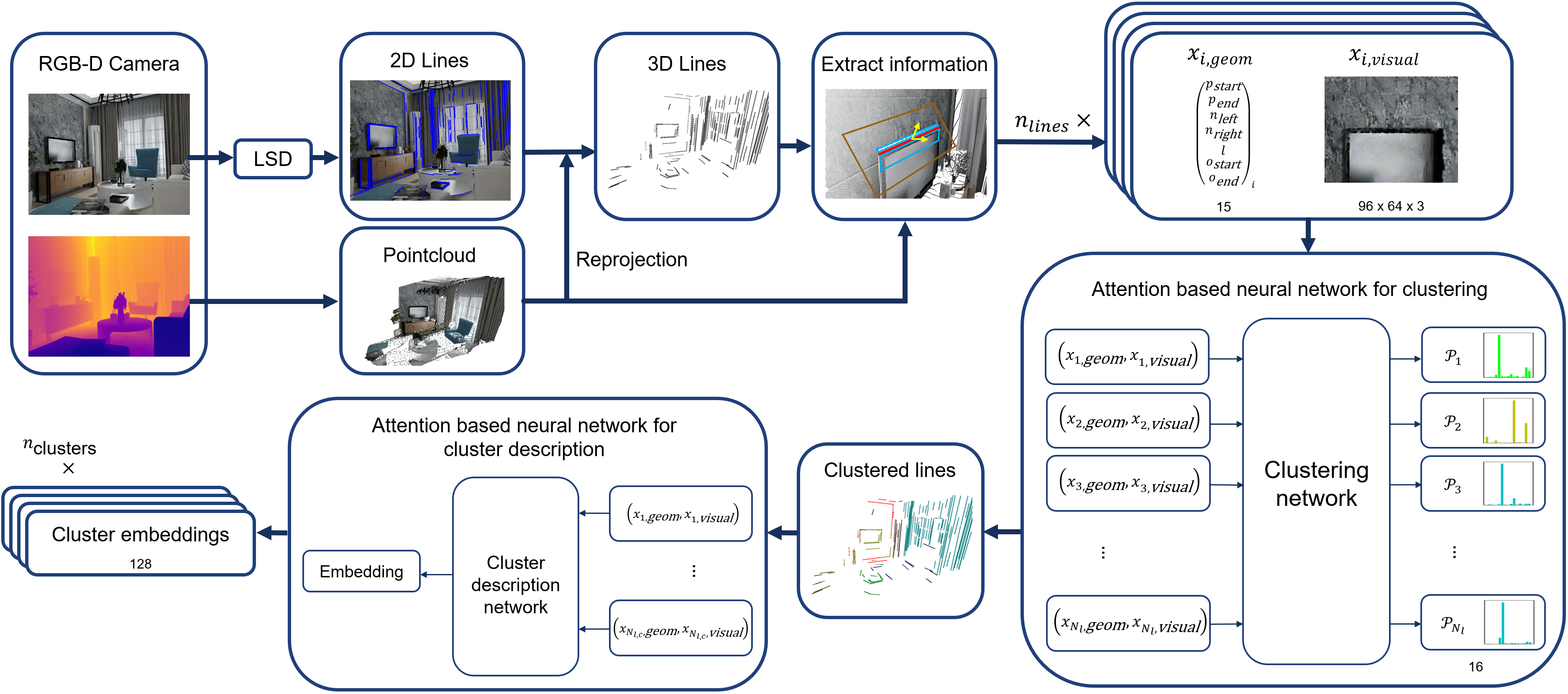}
        \caption{Summary of the proposed place recognition approach using line clusters as visual features.}
        \label{fig:pipeline}
    \end{center}
    \vspace{-0.6cm}
\end{figure*}

In our approach, we first detect line segments in an RGB camera image using the LSD line detector \cite{LSD}. Next, these lines are reprojected to 3D using the depth map. We capture additional geometric information and a virtual image of the environment around each line. The line geometries and virtual images are then processed by a line clustering neural network, which predicts a cluster label for each line. Each group of lines with equal cluster label is then processed by a second neural network for cluster description to obtain an embedding for each line cluster. Both networks are based on attention mechanisms. The cluster description embedding can be compared to cluster embeddings from other frames to recognize clusters and scenes. The complete pipeline is shown in Figure~\ref{fig:pipeline}.

\subsection{Extraction of line information}
\label{line_information}
In this section, we explain the inputs to the neural networks for clustering and cluster description and how they are generated from RGB-D images. 

\begin{figure}
    \centering
    \subfloat[Geometrical information]{{
        \includegraphics[width=0.4\linewidth]{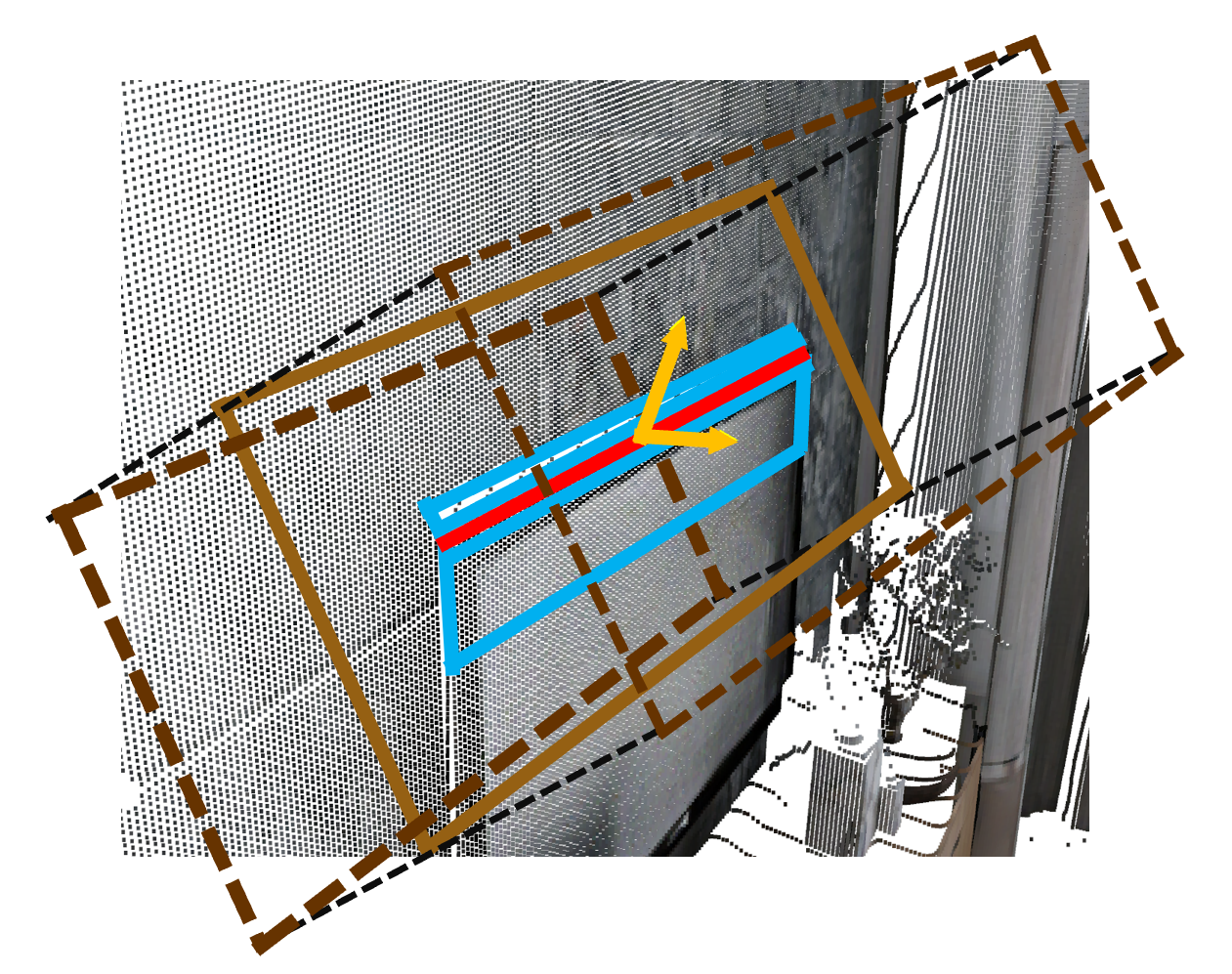}
        \label{subfig:geom}
    }}
    \quad
    \subfloat[Virtual image of environment]{{
        \includegraphics[width=0.4\linewidth]{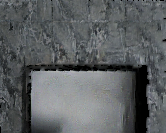}
        \label{subfig:vci}
    }}
    \caption{The geometrical information of a line together with a virtual camera image are used as the input to the neural network for clustering. The geometrical information~(a) consists of the normals (yellow) of the adjacent planes (blue) of the line (red), the line length and whether the line ends are visible. The brown box indicates the view frustum of the virtual camera, resulting in the virtual image in~(b).}
    \label{fig:line_geometry}
    \vspace{-0.6cm}
\end{figure}
\vspace{-0.3cm}
\subsubsection{Line detection and reprojection}
Line segments are extracted from gray-scaled RGB images using the LSD line detector \cite{LSD}, which is robust and fast. Co-linear line segments with close extremities are fused. Lines below a certain pixel length are rejected. 
Point clouds of the line neighborhood are obtained by pooling the depth map with rectangular regions on either side of the line. A plane is fitted on each of the neighborhood point clouds using RANSAC. If the planes are co-planar (texture line) or only one plane exists (\eg due to discontinuity), the line extremities are projected onto the corresponding plane. If the distance between the two plane centers along either of the surface normals is over a certain threshold ($\unit[0.3]{m}$), the line is projected onto the plane closer to the camera (discontinuity). Otherwise, the extremities are projected onto the intersection line of the two planes (edge).
\vspace{-0.3cm}
\subsubsection{Geometric information of line segment}
After extraction, information about the geometry of the line segments is recorded. It consists of the 3D start and end point $p_{s\text{/}e}$, and the normals $n_{l\text{/}r}$ of the previously described neighborhood planes (Figure~\ref{subfig:geom}). These planes give further structural information. If one plane does not exist, the corresponding normal is zero-padded. The line segment length $l$ is precomputed and appended to the information. Notice that the line length is already encoded in the extremities of the line, but it is added explicitly to help the neural networks to take length into account. Lastly, two boolean values $o_{s\text{/}e}$ are stored indicating whether the line extremities are occluded by an object in front of it or if they are on the edge of the view plane. The idea is that in case of occlusion, the actual position of extremities and length of the line is unknown and should thus be handled differently by the network. All geometrical information is stored in a 15-dimensional vector
\begin{equation}
    x_{i, \textit{geom}} = [ 
         p_{s},
         p_{e},
         n_{l},
         n_{r},
         l,
         o_{s},
         o_{e},
     ]^T_i \in 
    \left( \begin{array}{cc}
         \mathbb{R}^{13\times1}  \\
         \mathbb{B}^{2\times1}
    \end{array} \right).
\end{equation}
\vspace{-0.3cm}
\subsubsection{Virtual images of line environment}
To gather additional appearance information from the available RGB data, a viewpoint invariant virtual image is taken from the environment of the line. The point cloud in the vicinity of the line is orthogonal-projected onto a virtual camera plane. To make the perspective of this virtual camera invariant to the viewing angle, the camera plane is set at a fixed distance of 0.5\,m, parallel to the line and perpendicular to the mean of the two neighborhood plane normals. The camera has a maximum viewing distance of 1\,m and the field of view corresponds to 1.5 times of the line length in width, and one time in height (Figure~\ref{subfig:geom}).
The image is then discretized with a variable resolution to prevent aliasing. Despite variable resolution, these images suffer from scattered empty pixels. To deal with these holes, a dilated mask of non-black pixels is inpainted using the Navier-Stokes algorithm~\cite{NavierStokes} (Figure~\ref{subfig:vci}). Finally, it is linearly resized to a resolution of $96\times64$ and stored in the appearance information vector
\begin{equation}
    x_{i,\textit{visual}} \in \mathbb{R}^{96\times64\times3}.
\end{equation}

\subsection{Neural network for line clustering}

The task of the clustering network is to predict a cluster label for each line. The neural network needs to process each line, while simultaneously gathering information of the surroundings of the line, similar to what \acp{cnn} do with pixels. The idea is that lines need to ``know" their neighborhood lines to determine the correct correspondences. For example, the lines that make up a door frame are shaped like a door, and should therefore be assigned the same label. However, due to the unsortable nature of lines and the variable number of lines in each frame, \acp{cnn} or \acp{rnn} cannot be used without enforcing some kind of sorting. 

As a solution, scaled residual multi-head attention modules are used in the \textit{Transformer}~\cite{Attention}. It has been deployed successfully for natural language processing (NLP) tasks. The attention modules enable each word to ``communicate" with other words in the sentence to understand the context. In our work, these attention modules enable each line to gather information about its surrounding lines and the scene. 

The inputs of the network are the geometric and visual information of each line, and the output is a probability distribution, where each bin corresponds to the probability of a line belonging to a certain cluster. The predicted cluster label is then determined through \texttt{argmax}.
In the following, we describe in detail how the network is structured and trained, and how it achieves good clustering performance on indoor scenes.

\begin{figure*}
    \begin{center}
        \includegraphics[width=.8\linewidth]{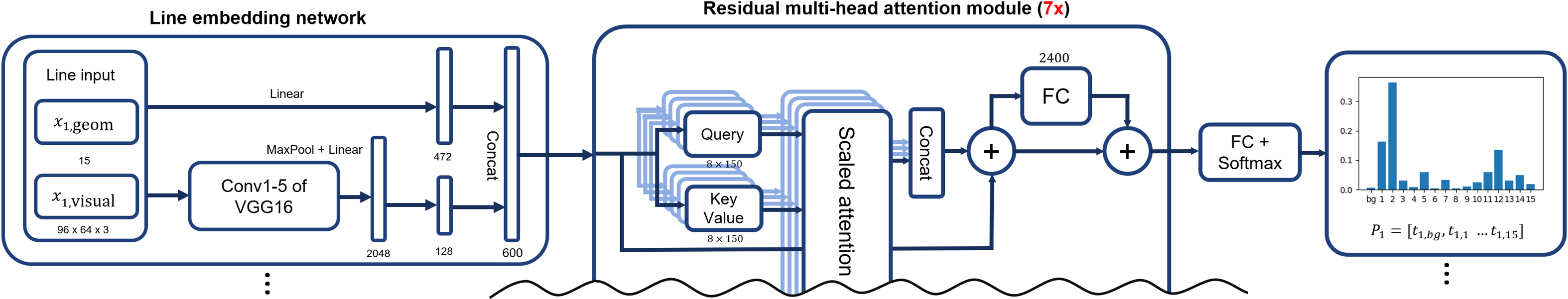}
        \caption{Part of the network architecture of the clustering network. The line embedding networks run in parallel for all lines. }
        \label{fig:clustering_network}
    \end{center}  
    \vspace{-0.6cm}
\end{figure*}
\vspace{-0.3cm}
\subsubsection{Network architecture}
\label{clustering_architecture}
The inputs of the network are the geometry and virtual image of each line $\{ \left( x_{i, \textit{geom}}, x_{i, \textit{visual}} \right) \}_{i=1}^{N_l}$, as described in Section \ref{line_information}. $N_l \leq N_{l,\textit{max}}$ is the total number of lines in a frame.

The geometric and visual data is preprocessed by the line embedding network. The geometric data is expanded with a linear layer and Leaky ReLu activation to form the $d_\textit{geom}=472$ dimensional geometric encoding. The virtual image is passed through the first 7 layers of the VGG-16 \ac{cnn} \cite{VGG}, $4\times4$ max-pooled and flattened. A hidden layer with $d_{h}=2048$ neurons is added to output a visual encoding with a dimension of $d_\textit{visual}=128$. The geometrical and visual encodings are concatenated to form line embeddings of dimension $d_\textit{model}=600$. The line embedding networks run in parallel for every line and share the same weights. Empty inputs are masked.

The main body of the network is composed of $N_\textit{layers} = 7$ identical layers adapted from the encoder modules from \textit{Transformer}~\cite{Attention}. Each layer is made up of two sub-layers, the scaled multi-head self-attention layer and the position-wise \ac{fc} feed-forward layer. 
Each sub-layer is bypassed with residual connections. Instead of normalization, Leaky ReLu activation with $\alpha=0.3$ is used as non-linearity. The residual output of each sub-layer is thus the Leaky ReLu activation of the sum of the input and the output of the sub-layer. In the attention sub-layer, four of the dot-product attention heads are replaced with additive attention heads, to handle the very diverse nature of the inputs. This results in $h_\textit{add}=4$ additive self-attention heads and $h_\textit{dot}=4$ dot-product self-attention heads per layer. The heads have a query/key dimensionality of $d_q=d_k=150$. The key and value tensors are identical. The inner layer of the feed-forward network has dimensionality $d_{ff}=2400$ and Leaky ReLu activation with $\alpha=0.3$. The dimension of all sub-layer outputs is $d_\textit{model}$ to facilitate the residual connections. Finally, a position-wise linear layer with a softmax activation outputs a probability distribution of dimension $d_\textit{label}=16$. Batch normalization is applied to all layer outputs to accelerate training. A graphical representation of the clustering network can be found in Figure~\ref{fig:clustering_network}.
\vspace{-0.3cm}
\subsubsection{Learning objective}
The learning objective used in this work is adapted from \cite{LearningToCluster}, which proposed a method for pixel-wise instance segmentation. Detailed motivation of the following losses can be found in their paper.

Let us denote the output of the clustering network as the probability distribution $\mathcal{P}_i = f(l_i)=[t_{i,\textit{bg}},t_{i,1}..t_{i,N_c}]$ where $N_c=d_\textit{label}-1$ indicates the number of instance clusters detectable in each frame. $t_{i, \textit{bg}}$ is the probability of line $l_i$ to belong to the background cluster, and $t_{i, k}$ is the probability of each line to belong to instance cluster $k$.

The task of the network is to assign a correct cluster label to each line. 
Lines of the same cluster should be assigned the same label, and lines of different clusters should be assigned different labels. We treat all background lines (\eg floor, ceiling, wall lines) specially, as they tend to be spread out across the room, and are not packed together like typical instances (\eg table, chair). The background label is absolute, and we reserve the first label $t_{i,0}$ for those and apply a binary cross-entropy loss for background classification over all lines
\begin{equation}
     \mathcal{L}_\textit{bg}=\sum_i^{N_l}(I_i^\textit{bg} \log t_{i,bg} + (1-I_i^\textit{bg}) \log (\sum_{k=1}^{N_c} t_{i,k})).
\end{equation}
Cluster labels of instances, however, are not absolute, since they only contain relative information. Switching labels of two clusters results in equivalent clustering. The label assignment is not unique. We define the relationship between any two non-background lines $l_i$ and $l_j$ as
\begin{equation}
    R(l_i,l_j)= 
    \begin{cases}
        1 & \text{if } l_i,l_j \text{ belong to the same instance,} \\
        0 & \text{otherwise.}
    \end{cases}
\end{equation}
We apply a KL-divergence based loss, which prefers similar probability distributions for lines of same labels, and dissimilar distributions for lines of different labels. For line pairs of the same label we apply a cost given by
\begin{multline}
\mathcal{L}(l_i,l_j)^+=\mathcal{D}_{KL}(\mathcal{P}_i^*||P_j)+\mathcal{D}_{KL}(\mathcal{P}_j^*||P_i), \\ 
\text{where } \mathcal{D}_{KL}(\mathcal{P}_i^*||P_j)=\sum_{k=1}^{N_c} t_{i,k} \log (\frac{t_{i,k}}{t_{j,k}}).
\end{multline}
For line pairs of different label we apply the hinge loss
\begin{multline}
    \mathcal{L} (l_i,l_j)^-=L_h(\mathcal{D}_{KL}(\mathcal{P}_i^*||P_j),\sigma)+L_h(\mathcal{D}_{KL}(\mathcal{P}_j^*||P_i),\sigma), \\
    \text{where } L_h(e, \sigma) = \max (0, \sigma - e).
\end{multline}
We use $\sigma=2$ as margin.
The contrastive loss is a combination of the above losses
\begin{multline}
    \mathcal{L} (l_i,l_j)=R(l_i,l_j) \mathcal{L} (l_i,l_j)^+ \\
    + (1-R(l_i,l_j)) \mathcal{L} (l_i,l_j)^-.
\end{multline}
Let $T=\{(l_i,l_j)\}_{ \forall i, j}$ contain all non-background line pairs, then the full pairwise loss is
\begin{equation}
    \mathcal{L}_\textit{pair}=\sum_{(l_i, l_j) \in T} \mathcal{L} (l_i,l_j).
\end{equation}
And finally we add the background and pairwise loss to formulate the full loss function for clustering
\begin{equation}
    \mathcal{L}_\textit{clustering}=\mathcal{L}_\textit{pair} + \mathcal{L}_\textit{bg}.
\end{equation}
\vspace{-0.3cm}
\subsubsection{Training}
\label{clustering_training}
The network is trained on the synthetic InteriorNet dataset \cite{InteriorNet18}. It consists of more than 10k scenes with 20 photo-realistic RGB-D frames each from random viewpoints. The camera sensor is modeled as a pinhole camera with perfect depth data. Since it is a synthetic dataset, it often reuses the same object models. To prevent overfitting to these object models, a subset of 3600 scenes is selected for training, and a subset of 400 scenes is selected for validation. 
The dataset contains ground truth semantic and instance label masks. The ground truth labelling of the lines is determined by majority voting of the pixels in the line neighborhood. 

The VGG layers of the line embedding network are initialized with ImageNet pretrained weights~\cite{ImageNet}. The visual encoding network is pretrained for 20 epochs using only visual data. During pretraining, the last VGG layers are gradually unfrozen until only the first two layers are left with the original ImageNet weights. The weights of the visual encoding network are again frozen during training of the full network. The full network is trained using the Adam optimizer~\cite{Adam} for 40 epochs, with an initial learning rate of $5e^{-5}$ and a gradual decay.
To limit memory usage during training, the maximum number of lines per frame is set to $N_{l,\textit{max}}=160$. During evaluation, the maximum number of lines per frame is set to $N_{l,\textit{max}}=220$. If a frame contains more than $N_{l,\textit{max}}$ lines, lines are sampled at random. 
Data augmentation is applied by rigidly rotating the frames and blacking out images at random.

Training took 6 days on an Nvidia RTX 2070 Super GPU. The average inference time for 220 lines is \unit[0.1]{s} with a memory consumption of $\unit[3.5]{GB}$. 

\subsubsection{Clustering performance}
\begin{figure}
    \centering
    \subfloat[Ground truth clustering]{{
        \includegraphics[width=0.35\linewidth]{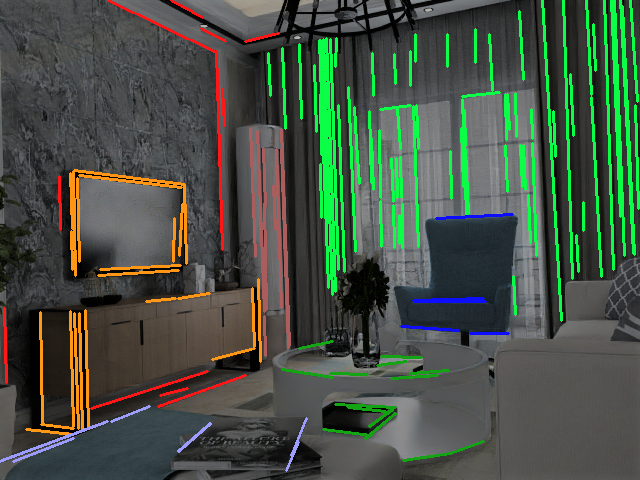}
    }}
    \quad
    \subfloat[Predicted clustering]{{
        \includegraphics[width=0.35\linewidth]{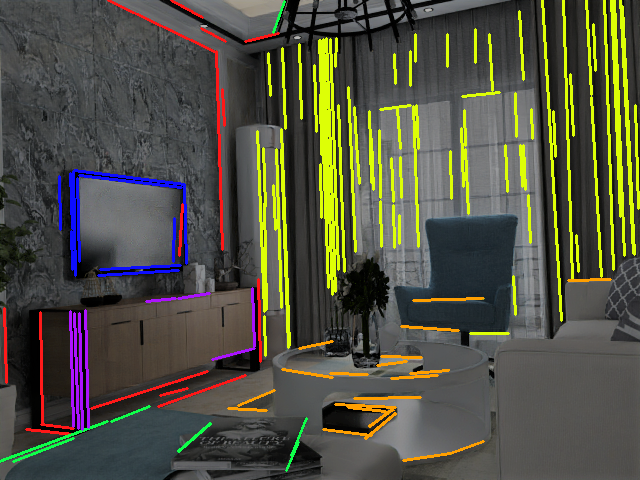}
    }}
    \vspace{-0.2cm}
    \caption{Comparison between \ac{gtc} and our prediction. Clusters with less than 4 lines are not shown. Different colors represent different cluster labels. Red lines are background lines. Note that absolute cluster colors are arbitrary without meaningful correspondences from (a) to (b).}
    \label{fig:clustering_example}
    \vspace{-0.6cm}
\end{figure}

Clustering performance is measured using the \ac{nmi} metric~\cite{NMI}. Background lines are treated as a separate cluster. It is compared to the conventional agglomerative clustering method \cite{Agglomerative}, where the minimum Euclidean distance between line segments is used as distance metric. The cluster count is determined using silhouette score minimization. To show that the attention layers indeed work, a vanilla \ac{fc} network (with the same architecture, with the exception of removed attention layers) is also trained and tested. The average \ac{nmi} over all frames of our validation subset of the InteriorNet dataset is determined for each method. 
Agglomerative clustering achieves an average \ac{nmi} of 0.45 while the vanilla \ac{fc} network improves on this with an average \ac{nmi} of 0.56. Using our clustering approach, we outperform both methods by achieving an average \ac{nmi} of 0.72 out of a perfect score of 1.
%
The agglomerative clustering approach mostly struggles to find the correct number of clusters, and fails to differentiate adjacent clusters. Qualitative results of our method can be seen in Figure~\ref{fig:clustering_example}.

\subsection{Neural network for cluster description}
Next, we explain how we describe line clusters with a descriptor embedding.
Line clusters are the groups of lines with the same predicted label. The maximum number of line clusters in each frame is thus $d_\textit{label}$. All groups of lines are sequentially fed into the network that uses the same inputs as the network for clustering. The network output is a descriptor embedding in $d_\textit{descriptor}=128$ dimensional euclidean space. 

\vspace{-0.3cm}
\subsubsection{Network architecture}
\begin{figure}
    \centering
    \includegraphics[width=.9\linewidth]{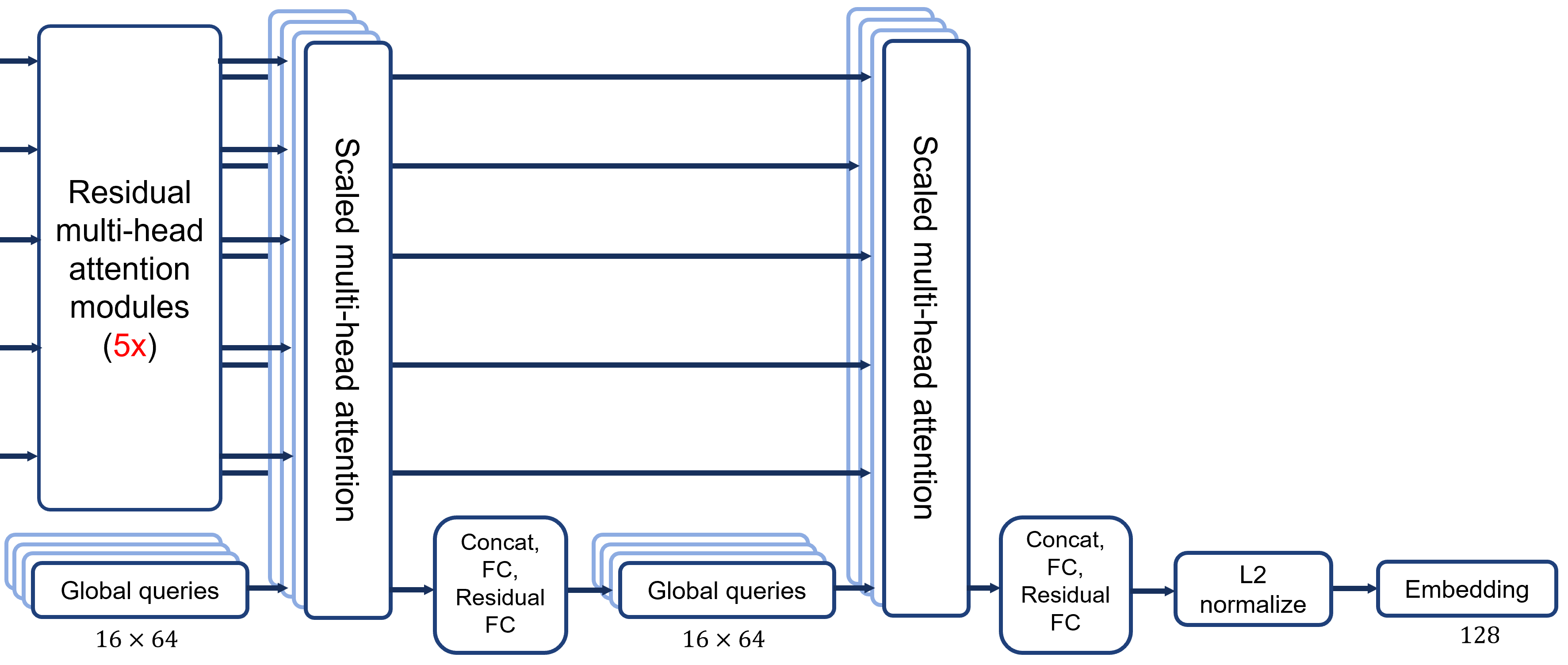}
    \caption{The tail part of the descriptor network consists of two scaled global multi-head attention layers gathering information over all lines using learned query vectors. This information is then concentrated into a normalized 128-dimensional embedding. }
    \label{fig:descriptor_network}
    \vspace{-0.3cm}
\end{figure}

The descriptor network uses the same line embedding network structure and the same residual multi-head attention layers as the clustering network. The dimensionality of the visual encoding is the same. Because the clusters contain less lines than frames and thus less structural information, the dimensionality of the geometric encoding is reduced to $d_\textit{geom}=384$, making the line embedding dimension $d_\textit{model}=512$. The dimensionality of the heads are reduced to $d_k=d_v=64$. The hidden layer of the residual feedforward modules has a dimension $d_{ff}=2048$. Furthermore, only $N_\textit{layers} = 5$ attention modules are used. At the output of the last attention module, two global attention modules are added (Figure~\ref{fig:descriptor_network}). They work in the same way as the previously shown self-attention modules. Each line produces key/value vectors, and learned global query vectors ``summarize" these values. The outputs of the attention modules are then concatenated. A linear layer with $d_\textit{global}=1024$ activations is added and the output is fed through a residual \ac{fc} layer. Linear layers then produce the global query vectors of the second global attention module. The outputs of this module are again concatenated and fed through a residual \ac{fc} network. A final linear layer and L2 normalization is added to produce the descriptor embedding that lies on a 128-dimensional hyper sphere. The hidden layers of each residual \ac{fc} layer have a dimension of $d_{fc}=4096$. Both global attention modules have $h_\textit{dot}=8$ and $h_\textit{add}=8$ heads with a dimensionality of $d_q=d_k=64$.

\vspace{-0.3cm}
\subsubsection{Triplet loss}
\begin{figure}
    \centering
    \subfloat[Anchor]{{
        \includegraphics[width=0.25\linewidth]{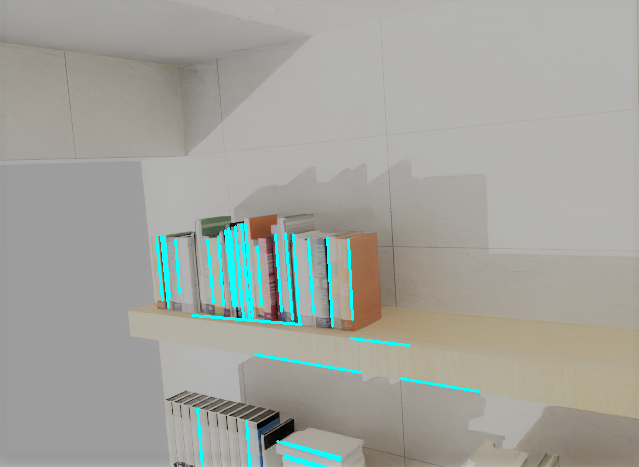}
    }}
    \subfloat[Positive]{{
        \includegraphics[width=0.25\linewidth]{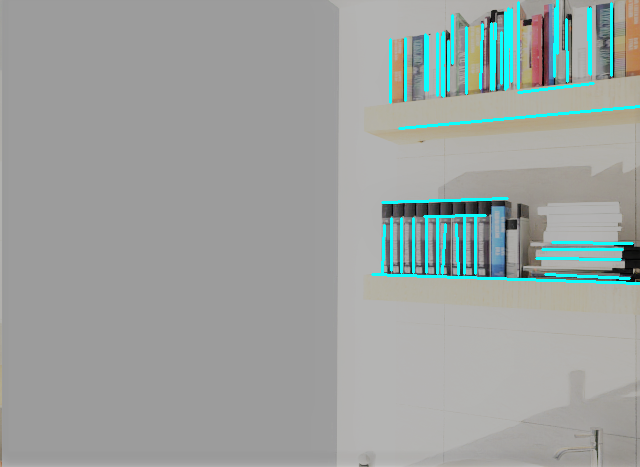}
    }}
    \subfloat[Negative]{{
        \includegraphics[width=0.25\linewidth]{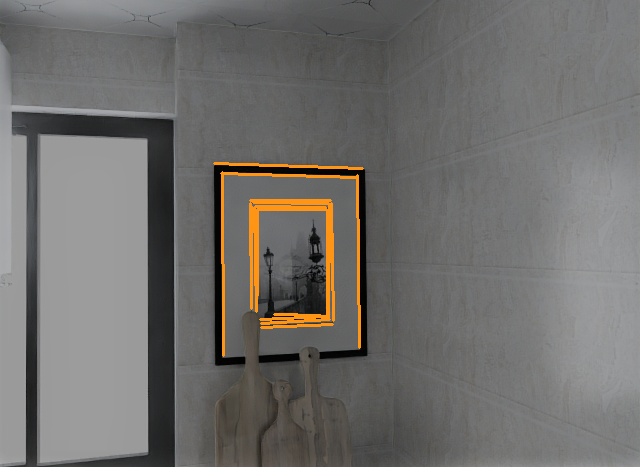}
    }}
    \vspace{-0.2cm}
    \caption{Example of a triplet used during training.}
    \label{fig:triplet}
    \vspace{-0.6cm}
\end{figure}

We use the triplet loss as described in~\cite{TripletLoss} for the object recognition learning task. We use triplets of clusters each consisting of an anchor, positive and negative cluster (Figure~\ref{fig:triplet}). The anchor and positive cluster are the same cluster viewed from another frame, and the negative cluster is a random different cluster. The objective is to increase the Euclidean distance between the anchor and negative cluster embeddings $y_{a}$ and $y_{n}$, while decreasing the Euclidean distance between anchor and positive cluster embeddings $y_{a}$ and $y_{p}$, respectively. The loss for each triplet in the training set is
\begin{equation}
\mathcal{L}_\textit{triplet} = \max (0, ||y_{a} - y_{p}||_2 - ||y_{a} - y_{n}||_2 + m).
\end{equation}
The margin $m$ is a hyper parameter and set to $0.6$.

\vspace{-0.3cm}
\subsubsection{Training}
Weights of the visual encoding network are taken from the clustering network and frozen during training.
We use the same subset of the InteriorNet dataset as described in Section~\ref{clustering_training}. In the dataset, each instance has the same label across all 20 frames of each scene. The positive cluster pairs are extracted by selecting an instance with more than one occurrence across these 20 frames as the anchor cluster, and the same instance from another view as positive cluster. The negative cluster is an instance selected at random from another scene. With a probability of $0.5$, the negative cluster is chosen to have the same semantic label as the anchor cluster. This is done to facilitate the differentiation of clusters of the same class. The network is trained using the Adam optimizer \cite{Adam} for 50 epochs, with initial learning rate of 0.0001 and a gradual decay. The same data augmentation is applied as described in Section~\ref{clustering_training}.

\subsection{Place recognition}
To perform place recognition, all RGB-D images from all scenes are passed through the proposed pipeline to obtain the line clusters and cluster embeddings for each frame. Clusters with less than 4 lines are removed to prevent matches based on insufficient information. Although omitted during the training of the cluster description network, the background is treated as any instance cluster during place recognition. The cluster embeddings of the map frames are assigned the corresponding scene index and stored in the map as a k-d tree for fast lookups. During query, the $k_{nn}$ \acp{nn} of each cluster embedding of the query frame are determined in the map. This results in $n_\textit{clusters} \times k_{nn}$ \acp{nn} for each frame, each assigned to a scene. The scene with most occurrences is selected and assigned to the query frame. 

\section{Evaluation}
We evaluate the performance of the full place recognition pipeline on three datasets, and compare it to three baselines. 


The performance of each method is measured by evaluating the ability to recognize scenes. The evaluation datasets consist of scenes with a number of frames each from different viewpoints. One frame is removed and the task is to assign this frame to the correct scene. This is repeated for every frame from every scene. The performance metric is the ratio of true positives (Accuracy), which is the ratio of correctly matched frames over the total number of frames.

We compare against a \ac{bow} approach using SIFT~\cite{sift}, and the learned \ac{sp}~\cite{SuperPoint} features. A vocabulary size of $d_\textit{sift}=512$ on the NYU and DIML datasets, and $d_\textit{sift}=800$ for the InteriorNet dataset is chosen for the SIFT $+$ \ac{bow} approach. For the \ac{sp} $+$ \ac{bow} method, we use the network weights from the original paper with a vocabulary size of $d_\textit{sp}=1024$. The vocabulary is trained on the corresponding evaluation dataset. In addition to this, we compare to the global descriptor NetVLAD~\cite{netvlad_tf} along with the best performing model and weights (VGG16 + whitening, trained on Pitts30k). The \ac{nn} frame is determined using the Euclidean distance between the vocabulary histograms. The query frame is then assigned to the scene of the \ac{nn} frame. 

\begin{figure}
    \centering
    \begin{tabular}{ccc}
        \includegraphics[height=0.18\linewidth]{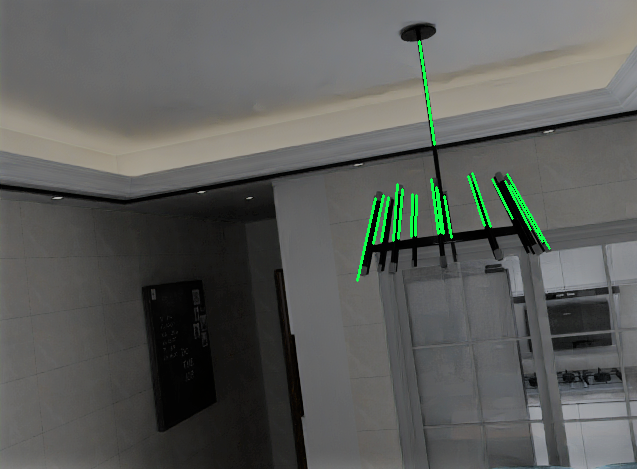}&
        \includegraphics[height=0.18\linewidth]{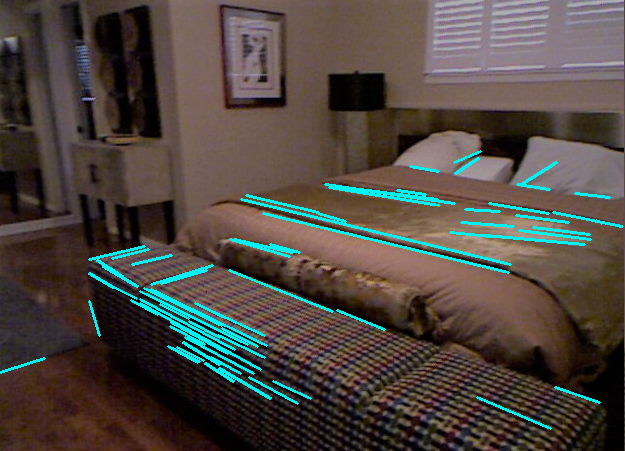}&
        \includegraphics[height=0.18\linewidth]{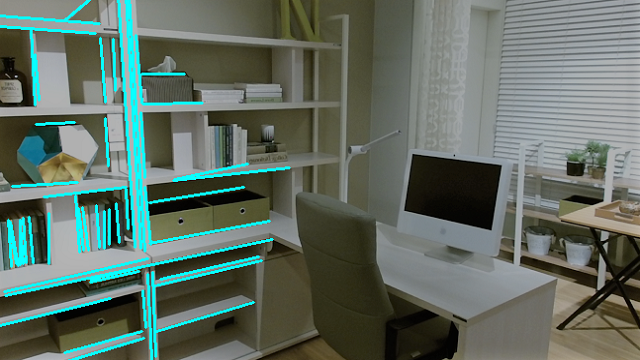}
        \\
        \includegraphics[height=0.18\linewidth]{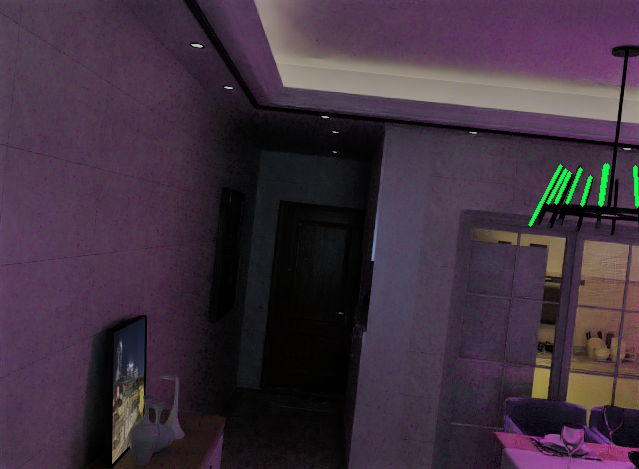}&
        \includegraphics[height=0.18\linewidth]{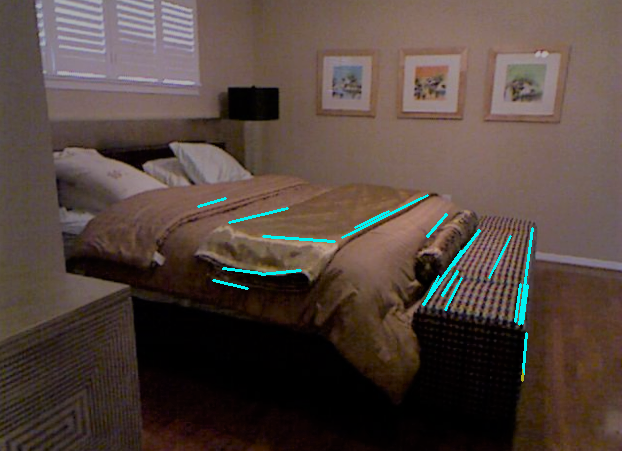}&
        \includegraphics[height=0.18\linewidth]{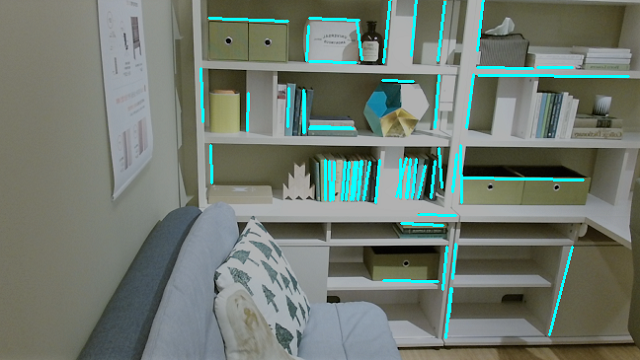}
        \\
        (a) InteriorNet&
        (b) NYU&
        (c) DIML
    \end{tabular}
    \vspace{-0.2cm}
    \caption{Examples of frames where the SIFT $+$ \ac{bow} approach failed. Our method could recognize the scene by recognizing the objects, even under large viewpoint and lighting changes. The matched line clusters are indicated with colored lines.}
    \label{fig:matches}
    \vspace{-0.3cm}
\end{figure}

\subsection{InteriorNet dataset}
The InteriorNet subset used for evaluation is the validation set used during training (Section~\ref{clustering_training}). It contains 400 scenes, consisting of 20 frames captured in a single room. However, some of these scenes are connected, because some rooms are located on the same floor. We treat these rooms as the same scene during evaluation, resulting in 100 separate scenes. Large viewpoint changes are present because of these connected rooms. The number of \acp{nn} chosen for our approach is $k_{nn}=8$. Since ground truth instance masks are available, our pipeline is also tested using \ac{gtc}. That is, the neural network for clustering is replaced with ground truth cluster labels. Our approach outperforms the SIFT $+$ \ac{bow} baseline by $\unit{52}{\%}$ and the more recent \ac{sp} $+$ \ac{bow} by $\unit{16}{\%}$, as depicted in Table~\ref{tab:all_results}. The indoor environments of the InteriorNet dataset are very clean and consistent, thus local features are very repetitive (corners of different doors, drawers, tables, etc. appear the same), making  the advantage of using high-level structural features apparent. The predicted line clusters are not as consistent between frames as the ground truth clusters, therefore the performance with \ac{gtc} is improved. However, the strong performance of our full pipeline shows that the descriptor is robust to missing and wrongly assigned lines.

The InteriorNet dataset contains randomly illuminated versions of all frames. We tested the methods on illumination invariance by replacing the query frames with randomly illuminated frames (INet + RI), i.e. the map frames are regularly illuminated and the query frames are randomly illuminated. For this evaluation, we retrained our descriptor and cluster model and included frames containing random lighting during training. Our pipeline performs better than the methods using local features (SIFT/\ac{sp} $+$ \ac{bow}), demonstrating the advantage of using line clusters under changing lighting. 

NetVLAD significantly outperforms all other methods in all experiments. It benefits from the unique style of each scene designed by interior architects and can thus match frames even with no viewpoint overlap. Also, the dimensionality of the NetVLAD descriptor ($4096$) is significantly larger than ours (approx. $800$ per frame).

\begin{table}[h]
\begin{center}
\begin{tabular}{@{}ccccc@{}}
\toprule
\multirow{2}{*}{Method} &  \multicolumn{4}{c}{Accuracy}\\
\cmidrule{2-5}
  & InteriorNet & INet + RI & DIML & NYU \\
\midrule
SIFT + BoW & 38\% & 21\% & 66\% & 30\% \\
SP + BoW & 50\% & 32\% & 73 \% & 26\% \\
NetVLAD & 93\% & 91\% & 94\% & 84\% \\
\midrule
Ours & 58\% & 34\% & 65\% & 20\% \\
Ours, GTC & 64\% & 41\% & n/a & n/a\\
\bottomrule
\end{tabular}
\end{center}
\vspace{-0.3cm}
\caption{Evaluation on the synthetic InteriorNet dataset, the InteriorNet dataset with random illumination (INet + RI), the DIML dataset and the NYU dataset. Our approach performs significantly better than the SIFT $+$ \ac{bow} approach on the InteriorNet dataset. Using perfect \ac{gtc}, even better performance could be achieved. Results are slightly inferior on the real datasets because of depth noise and clutter. }
\label{tab:all_results}
\vspace{-0.3cm}
\end{table}

\subsection{DIML dataset}

The DIML dataset~\cite{diml} consists of a large number of real indoor and outdoor scenes. We use only indoor scenes, and pick only scene types that appear in the InteriorNet dataset, such as bedrooms, living rooms and kitchens. Our selection consists of 30 scenes with 8 frames each. We set $k_{nn}=4$. 
The scenes of the DIML dataset are captured using the KinectV2 RGB-D sensor. The depth data is reprojected into the RGB camera frame and post-processed to fill any depth holes resulting from reprojection or missing data. The noisy depth data causes slightly offset line geometry and noisy virtual camera images. However, our pipeline still achieves similar results to the SIFT $+$ \ac{bow} approach shown in Table \ref{tab:all_results}, despite being trained exclusively on a synthetic dataset with perfect depth data. Our approach could recognize 26 frames that could not be recognized by the SIFT approach (Figure~\ref{fig:matches}). The accuracy on the DIML dataset is higher than on the InteriorNet dataset for all approaches, because the number of scenes and frames is smaller.

\subsection{NYU dataset}
The NYU Depth Dataset V2~\cite{NYU} is also comprised of a variety of indoor scenes. The RGB-D images were obtained using the Microsoft Kinect V1 sensor, and missing depth data is filled using a colorization scheme. We use the labelled subset, and again select only scenes of types that appear in the InteriorNet dataset and contain 4 or more frames. The resulting selection consists of 52 scenes and 290 frames. We set $k_{nn}=8$.
Because this dataset is captured using an older version of the Kinect sensor, the depth and RGB data is less accurate and more noisy. Besides, the scenes are significantly more cluttered than the DIML dataset, which our approach is not trained to handle. These properties of the NYU dataset are reflected in the results shown in Table~\ref{tab:all_results}, where our approach performs slightly inferior to the SIFT $+$ \ac{bow} and SP $+$ \ac{bow} approach. Nevertheless, there are 35 frames recognized by our approach, but not by the SIFT $+$ \ac{bow} approach, an example is shown in Figure~\ref{fig:matches}. 

\section{Limitations and Outlook}
Our approach typically works better with large viewpoint changes and lighting changes compared to SIFT $+$ \ac{bow} and in some cases better than \ac{sp} $+$ \ac{bow}. It also handles clean and tidy rooms better. However, clutter adds numerous lines that impact the performance of the clustering network negatively. Therefore, in scenes with a lot of clutter, local keypoint approaches work better, as there are more keypoints provided there is no significant viewpoint change. Some DIML scenes and especially the scenes of the NYU dataset are very cluttered, explaining the weaker performance of our method. Furthermore, all frames of a scene are captured inside a single room for both real datasets, making the viewpoint changes limited. 
While our approach clearly outperforms the local feature based baseline approaches on simulated data, the performance decreases on real data. The cause of this decrease can be explained by having sparser and more corrupt virtual images as real world depth data has many holes and noise.
A large portion of this issue could be solved directly by using more recent sensors that provide denser and more accurate depth images, such as the Kinect Azure. In addition, sequential depth frames can be fused to generate more complete virtual images. Rendering techniques that are more robust to empty pixels, such as surface generation or voxel based rendering, can be used as well.

Our place recognition performance can also be improved by adding other types of visual features such as keypoints, curves and ellipses thanks to the flexibility of  our network architecture.
Finally, the matched line clusters can be used to perform 6-DoF pose estimation and matches can be improved through geometric consistency.

\section{Conclusion}
We introduced a novel learning-based approach that uses line clusters for place recognition. While outperforming conventional image-retrieval methods, we show comparable performance to learning-based local features descriptors. Even though, NetVLAD performs significantly better in recognizing a scene, our approach has the potential to be used in a 6-DoF localization pipeline. We showed how attention mechanism based neural networks can be used to both cluster lines and describe line clusters in RGB-D frames. These networks have a huge potential to also be trained and used for other clustering and description tasks. 
Additionally, we demonstrated a novel technique to describe lines with virtual camera images. This method improves viewpoint invariance and can be interesting for future research.


\begin{acronym}
\acro{nmi}[NMI]{Normalized Mutual Information}
\acro{cnn}[CNN]{Convolutional Neural Network}
\acro{rnn}[RNN]{Recurrent Neural Network}
\acro{bow}[BoW]{Bag-of-Words}
\acro{gtc}[GTC]{Ground Truth Clustering}
\acro{slam}[SLAM]{Simultaneous Localization And Mapping}
\acro{dl}[DL]{Deep Learning}
\acro{sp}[SP]{SuperPoint}
\acro{fc}[FC]{fully connected}
\acro{nn}[NN]{Nearest Neighbor}
\end{acronym}
{\small
\bibliographystyle{ieee}
\bibliography{egbib}

\begin{thebibliography}{10}\itemsep=-1pt

\bibitem{edlines}
C.~Akinlar and C.~Topal.
\newblock Edlines: Real-time line segment detection by edge drawing (ed).
\newblock pages 2837--2840, 09 2011.

\bibitem{IncrementalSLAM}
A.~{Angeli}, S.~{Doncieux}, J.~{Meyer}, and D.~{Filliat}.
\newblock Incremental vision-based topological slam.
\newblock In {\em 2008 IEEE/RSJ International Conference on Intelligent Robots
  and Systems}, pages 1031--1036, 2008.

\bibitem{netvlad}
R.~Arandjelovic, P.~Gron{\'{a}}t, A.~Torii, T.~Pajdla, and J.~Sivic.
\newblock Netvlad: {CNN} architecture for weakly supervised place recognition.
\newblock {\em CoRR}, abs/1511.07247, 2015.

\bibitem{arandjelovic2013vlad}
R.~{Arandjelovic} and A.~{Zisserman}.
\newblock All about vlad.
\newblock In {\em 2013 IEEE Conference on Computer Vision and Pattern
  Recognition}, pages 1578--1585, 2013.

\bibitem{ImproveObjectRetrieval}
R.~{Arandjelović} and A.~{Zisserman}.
\newblock Three things everyone should know to improve object retrieval.
\newblock In {\em 2012 IEEE Conference on Computer Vision and Pattern
  Recognition}, pages 2911--2918, 2012.

\bibitem{Surf}
H.~Bay, T.~Tuytelaars, and L.~Van~Gool.
\newblock Surf: Speeded up robust features.
\newblock In A.~Leonardis, H.~Bischof, and A.~Pinz, editors, {\em Computer
  Vision -- ECCV 2006}, pages 404--417, Berlin, Heidelberg, 2006. Springer
  Berlin Heidelberg.

\bibitem{NavierStokes}
M.~Bertalmio, A.~Bertozzi, and G.~Sapiro.
\newblock Navier-stokes, fluid dynamics, and image and video inpainting.
\newblock volume~1, pages I--355, 02 2001.

\bibitem{StereoSequences}
C.~{Cadena}, D.~{Galvez-López}, J.~D. {Tardos}, and J.~{Neira}.
\newblock Robust place recognition with stereo sequences.
\newblock {\em IEEE Transactions on Robotics}, 28(4):871--885, 2012.

\bibitem{diml}
J.~Cho, D.~Min, Y.~Kim, and K.~Sohn.
\newblock A large {RGB-D} dataset for semi-supervised monocular depth
  estimation.
\newblock {\em CoRR}, abs/1904.10230, 2019.

\bibitem{netvlad_tf}
T.~Cieslewski, S.~Choudhary, and D.~Scaramuzza.
\newblock Data-efficient decentralized visual {SLAM}.
\newblock {\em CoRR}, abs/1710.05772, 2017.

\bibitem{FABMAP}
M.~Cummins and P.~Newman.
\newblock Fab-map: Probabilistic localization and mapping in the space of
  appearance.
\newblock {\em I. J. Robotic Res.}, 27:647--665, 06 2008.

\bibitem{FABMAP2_0}
M.~Cummins and P.~Newman.
\newblock Highly scalable appearance-only slam - fab-map 2.0.
\newblock 06 2009.

\bibitem{Agglomerative}
D.~Defays.
\newblock {An efficient algorithm for a complete link method}.
\newblock {\em The Computer Journal}, 20(4):364--366, 01 1977.

\bibitem{SuperPoint}
D.~DeTone, T.~Malisiewicz, and A.~Rabinovich.
\newblock Superpoint: Self-supervised interest point detection and description.
\newblock {\em CoRR}, abs/1712.07629, 2017.

\bibitem{PhysicalWords}
R.~Finman, T.~Whelan, L.~Paull, and J.~J. Leonard.
\newblock Physical words for place recognition in dense {RGB-D} maps.
\newblock In {\em ICRA workshop on visual place recognition in changing
  environments}, 05/2014 2014.

\bibitem{furrer2020phd_thesis}
F.~Furrer.
\newblock {\em Online Incremental Object-Based Mapping for Mobile
  Manipulation}.
\newblock PhD thesis, ETH Zurich, Zurich, 2020-05.

\bibitem{ObjectDatabase}
F.~{Furrer}, T.~{Novkovic}, M.~{Fehr}, A.~{Gawel}, M.~{Grinvald}, T.~{Sattler},
  R.~{Siegwart}, and J.~{Nieto}.
\newblock Incremental object database: Building 3d models from multiple partial
  observations.
\newblock In {\em 2018 IEEE/RSJ International Conference on Intelligent Robots
  and Systems (IROS)}, pages 6835--6842, 2018.

\bibitem{Lost}
S.~Garg, N.~S{\"{u}}nderhauf, and M.~Milford.
\newblock Lost? appearance-invariant place recognition for opposite viewpoints
  using visual semantics.
\newblock {\em CoRR}, abs/1804.05526, 2018.

\bibitem{LSD}
R.~Gioi, J.~Jakubowicz, J.-M. Morel, and G.~Randall.
\newblock Lsd: A fast line segment detector with a false detection control.
\newblock {\em IEEE transactions on pattern analysis and machine intelligence},
  32:722--32, 04 2010.

\bibitem{Gojcic_2019_CVPR}
Z.~Gojcic, C.~Zhou, J.~D. Wegner, and A.~Wieser.
\newblock The perfect match: 3d point cloud matching with smoothed densities.
\newblock In {\em Proceedings of the IEEE/CVF Conference on Computer Vision and
  Pattern Recognition (CVPR)}, June 2019.

\bibitem{SceneSequences}
K.~Ho and P.~Newman.
\newblock Detecting loop closure with scene sequences.
\newblock {\em International Journal of Computer Vision}, 74:261--286, 07 2007.

\bibitem{LearningToCluster}
Y.~Hsu, Z.~Xu, Z.~Kira, and J.~Huang.
\newblock Learning to cluster for proposal-free instance segmentation.
\newblock {\em CoRR}, abs/1803.06459, 2018.

\bibitem{6631111}
{Jin Han Lee}, G.~{Zhang}, J.~{Lim}, and I.~H. {Suh}.
\newblock Place recognition using straight lines for vision-based slam.
\newblock In {\em 2013 IEEE International Conference on Robotics and
  Automation}, pages 3799--3806, 2013.

\bibitem{ContentBasedRetrieval}
{Junqiu Wang}, {Hongbin Zha}, and R.~{Cipolla}.
\newblock Combining interest points and edges for content-based image
  retrieval.
\newblock In {\em IEEE International Conference on Image Processing 2005},
  volume~3, pages III--1256, 2005.

\bibitem{Adam}
D.~Kingma and J.~Ba.
\newblock Adam: A method for stochastic optimization.
\newblock {\em International Conference on Learning Representations}, 12 2014.

\bibitem{DLD_line}
M.~Lange, F.~Schweinfurth, and A.~Schilling.
\newblock Dld: A deep learning based line descriptor for line feature matching.
\newblock {\em 2019 IEEE/RSJ International Conference on Intelligent Robots and
  Systems (IROS)}, pages 5910--5915, 2019.

\bibitem{OutdoorStraightLines}
J.~H. {Lee}, S.~{Lee}, G.~{Zhang}, J.~{Lim}, W.~K. {Chung}, and I.~H. {Suh}.
\newblock Outdoor place recognition in urban environments using straight lines.
\newblock In {\em 2014 IEEE International Conference on Robotics and Automation
  (ICRA)}, pages 5550--5557, 2014.

\bibitem{InteriorNet18}
W.~Li, S.~Saeedi, J.~McCormac, R.~Clark, D.~Tzoumanikas, Q.~Ye, Y.~Huang,
  R.~Tang, and S.~Leutenegger.
\newblock Interiornet: Mega-scale multi-sensor photo-realistic indoor scenes
  dataset.
\newblock In {\em British Machine Vision Conference (BMVC)}, 2018.

\bibitem{sift}
D.~G. {Lowe}.
\newblock Object recognition from local scale-invariant features.
\newblock In {\em Proceedings of the Seventh IEEE International Conference on
  Computer Vision}, volume~2, pages 1150--1157 vol.2, 1999.

\bibitem{PlaceRecognitionSummary}
S.~Lowry, N.~Sünderhauf, P.~Newman, J.~Leonard, D.~Cox, P.~Corke, and
  M.~Milford.
\newblock Visual place recognition: A survey.
\newblock {\em IEEE Transactions on Robotics}, pages 1--19, 11 2015.

\bibitem{RobustPointAndLines1}
Y.~{Lu} and D.~{Song}.
\newblock Robust rgb-d odometry using point and line features.
\newblock In {\em 2015 IEEE International Conference on Computer Vision
  (ICCV)}, pages 3934--3942, 2015.

\bibitem{RobustPointAndLines2}
Y.~{Lu} and D.~{Song}.
\newblock Robustness to lighting variations: An rgb-d indoor visual odometry
  using line segments.
\newblock In {\em 2015 IEEE/RSJ International Conference on Intelligent Robots
  and Systems (IROS)}, pages 688--694, 2015.

\bibitem{ma2020image}
J.~Ma, X.~Jiang, A.~Fan, J.~Jiang, and J.~Yan.
\newblock Image matching from handcrafted to deep features: A survey.
\newblock {\em International Journal of Computer Vision}, pages 1--57, 2020.

\bibitem{7298936}
B.~{Micusik} and H.~{Wildenauer}.
\newblock Descriptor free visual indoor localization with line segments.
\newblock In {\em 2015 IEEE Conference on Computer Vision and Pattern
  Recognition (CVPR)}, pages 3165--3173, 2015.

\bibitem{NYU}
P.~K. Nathan~Silberman, Derek~Hoiem and R.~Fergus.
\newblock Indoor segmentation and support inference from rgbd images.
\newblock In {\em ECCV}, 2012.

\bibitem{Gist}
A.~Oliva and A.~Torralba.
\newblock Torralba, a.: Building the gist of a scene: The role of global image
  features in recognition. progress in brain research 155, 23-36.
\newblock {\em Progress in brain research}, 155:23--36, 02 2006.

\bibitem{FABMAP3D}
R.~{Paul} and P.~{Newman}.
\newblock Fab-map 3d: Topological mapping with spatial and visual appearance.
\newblock In {\em 2010 IEEE International Conference on Robotics and
  Automation}, pages 2649--2656, 2010.

\bibitem{ObjectRetrieval}
J.~{Philbin}, O.~{Chum}, M.~{Isard}, J.~{Sivic}, and A.~{Zisserman}.
\newblock Object retrieval with large vocabularies and fast spatial matching.
\newblock In {\em 2007 IEEE Conference on Computer Vision and Pattern
  Recognition}, pages 1--8, 2007.

\bibitem{LineClusteringLane}
F.~{Poggenhans}, A.~{Hellmund}, and C.~{Stiller}.
\newblock Application of line clustering algorithms for improving road feature
  detection.
\newblock In {\em 2016 IEEE 19th International Conference on Intelligent
  Transportation Systems (ITSC)}, pages 2456--2461, 2016.

\bibitem{ImageNet}
O.~Russakovsky, J.~Deng, H.~Su, J.~Krause, S.~Satheesh, S.~Ma, Z.~Huang,
  A.~Karpathy, A.~Khosla, M.~S. Bernstein, A.~C. Berg, and F.~Li.
\newblock Imagenet large scale visual recognition challenge.
\newblock {\em CoRR}, abs/1409.0575, 2014.

\bibitem{sarlin2019coarse}
P.-E. Sarlin, C.~Cadena, R.~Siegwart, and M.~Dymczyk.
\newblock From coarse to fine: Robust hierarchical localization at large scale.
\newblock In {\em Proceedings of the IEEE Conference on Computer Vision and
  Pattern Recognition}, pages 12716--12725, 2019.

\bibitem{3DSift}
P.~Scovanner, S.~Ali, and M.~Shah.
\newblock A 3-dimensional sift descriptor and its application to action
  recognition.
\newblock pages 357--360, 01 2007.

\bibitem{VGG}
K.~Simonyan and A.~Zisserman.
\newblock Very deep convolutional networks for large-scale image recognition.
\newblock {\em arXiv 1409.1556}, 09 2014.

\bibitem{BagOfWords}
{Sivic} and {Zisserman}.
\newblock Video google: a text retrieval approach to object matching in videos.
\newblock In {\em ICCV 2003}, volume~9, pages 1470--1477 vol.2, 2003.

\bibitem{spezialetti20203d}
R.~Spezialetti, S.~Salti, L.~Di~Stefano, and F.~Tombari.
\newblock 3d local descriptors—from handcrafted to learned.
\newblock {\em 3D Imaging, Analysis and Applications}, pages 319--352, 2020.

\bibitem{svarm2016city}
L.~Sv{\"a}rm, O.~Enqvist, F.~Kahl, and M.~Oskarsson.
\newblock City-scale localization for cameras with known vertical direction.
\newblock {\em IEEE transactions on pattern analysis and machine intelligence},
  39(7):1455--1461, 2016.

\bibitem{ConvNetLandmarks}
N.~Sünderhauf, S.~Shirazi, A.~Jacobson, E.~Pepperell, F.~Dayoub, B.~Upcroft,
  and M.~Milford.
\newblock Place recognition with convnet landmarks: Viewpoint-robust,
  condition-robust, training-free.
\newblock 07 2015.

\bibitem{Tang2019place}
X.~Tang, W.~Fu, M.~Jiang, G.~Peng, Z.~Wu, Y.~Yue, and D.~Wang.
\newblock Place recognition using line-junction-lines in urban environments.
\newblock pages 530--535, 11 2019.

\bibitem{toft2018semantic}
C.~Toft, E.~Stenborg, L.~Hammarstrand, L.~Brynte, M.~Pollefeys, T.~Sattler, and
  F.~Kahl.
\newblock Semantic match consistency for long-term visual localization.
\newblock In {\em Proceedings of the European Conference on Computer Vision
  (ECCV)}, pages 383--399, 2018.

\bibitem{SiftSurfSeasons}
C.~Valgren and A.~Lilienthal.
\newblock Sift, surf and seasons: Long-term outdoor localization using local
  features.
\newblock 09 2007.

\bibitem{Attention}
A.~Vaswani, N.~Shazeer, N.~Parmar, J.~Uszkoreit, L.~Jones, A.~N. Gomez,
  L.~Kaiser, and I.~Polosukhin.
\newblock Attention is all you need.
\newblock {\em CoRR}, abs/1706.03762, 2017.

\bibitem{verhagen2014scale}
B.~Verhagen, R.~Timofte, and L.~Van~Gool.
\newblock Scale-invariant line descriptors for wide baseline matching.
\newblock In {\em IEEE Winter Conference on Applications of Computer Vision},
  pages 493--500. IEEE, 2014.

\bibitem{NMI}
N.~X. Vinh, J.~Epps, and J.~Bailey.
\newblock Information theoretic measures for clusterings comparison: Variants,
  properties, normalization and correction for chance.
\newblock {\em J. Mach. Learn. Res.}, 11:2837–2854, Dec. 2010.

\bibitem{wang2009wide}
L.~Wang, U.~Neumann, and S.~You.
\newblock Wide-baseline image matching using line signatures.
\newblock In {\em 2009 IEEE 12th International Conference on Computer Vision},
  pages 1311--1318. IEEE, 2009.

\bibitem{MSLD_line}
Z.~Wang, F.~Wu, and Z.~hu.
\newblock Msld: A robust descriptor for line matching.
\newblock {\em Pattern Recognition}, 42:941--953, 05 2009.

\bibitem{weyand2016planet}
T.~Weyand, I.~Kostrikov, and J.~Philbin.
\newblock Planet-photo geolocation with convolutional neural networks.
\newblock In {\em European Conference on Computer Vision}, pages 37--55.
  Springer, 2016.

\bibitem{TripletLoss}
P.~Wohlhart and V.~Lepetit.
\newblock Learning descriptors for object recognition and 3d pose estimation.
\newblock {\em CoRR}, abs/1502.05908, 2015.

\bibitem{LocalDeepFeatures}
A.~B. {Yandex} and V.~{Lempitsky}.
\newblock Aggregating local deep features for image retrieval.
\newblock In {\em 2015 IEEE International Conference on Computer Vision
  (ICCV)}, pages 1269--1277, 2015.

\bibitem{3Dmatch}
A.~Zeng, S.~Song, M.~Nie{\ss}ner, M.~Fisher, J.~Xiao, and T.~Funkhouser.
\newblock 3dmatch: Learning local geometric descriptors from rgb-d
  reconstructions.
\newblock In {\em CVPR}, 2017.

\bibitem{3DLineBasedSLAM}
G.~{Zhang}, J.~H. {Lee}, J.~{Lim}, and I.~H. {Suh}.
\newblock Building a 3-d line-based map using stereo slam.
\newblock {\em IEEE Transactions on Robotics}, 31(6):1364--1377, 2015.

\bibitem{zhang2013efficient}
L.~Zhang and R.~Koch.
\newblock An efficient and robust line segment matching approach based on lbd
  descriptor and pairwise geometric consistency.
\newblock {\em Journal of Visual Communication and Image Representation},
  24(7):794--805, 2013.

\end{thebibliography}
}

\end{document}